# The Complexity of Plan Existence and Evaluation in Probabilistic Domains

**Judy Goldsmith**
Dept. of Computer Science
University of Kentucky
Lexington, KY 40506-0046
goldsmit@cs.uky.edu

**Michael L. Littman**
Dept. of Computer Science
Duke University
Durham, NC 27708-0129
mlittman@cs.duke.edu

**Martin Mundhenk**
FB4 - Theoretische Informatik
Universität Trier
D-54286 Trier
Germany
mundhenk@ti.uni-trier.de

## Abstract

We examine the computational complexity of testing and finding small plans in probabilistic planning domains with succinct representations. We find that many problems of interest are complete for a variety of complexity classes: NP, co-NP, PP, NP$^{PP}$, co-NP$^{PP}$, and PSPACE. Of these, the probabilistic classes PP and NP$^{PP}$ are likely to be of special interest in the field of uncertainty in artificial intelligence and are deserving of additional study. These results suggest a fruitful direction of future algorithmic development.

## 1 INTRODUCTION

Recent work in artificial-intelligence planning has addressed the problem of finding effective plans in domains in which operators have probabilistic effects (Kushmerick, Hanks, & Weld 1995; Draper, Hanks, & Weld 1994; Dearden & Boutilier 1997; Boutilier, Dearden, & Goldszmidt 1995; Boutilier, Dean, & Hanks 1995). In *probabilistic propositional planning*, operators are specified in a Bayes network or an extended STRIPS-like notation, and the planner is asked to determine a way of choosing operators to achieve a goal configuration with some user-specified probability. This problem is closely related to that of solving a Markov decision process when it is expressed in a succinct representation.

In previous work (Littman 1997; Mundhenk, Goldsmith, & Allender 1996), we examined the complexity of determining whether a valid plan exists; the problem is EXP-complete in its general form and PSPACE-complete when we are limited to polynomial-depth plans. For these results to hold, plans must be permitted to be arbitrarily complicated objects, and there is no restriction that a valid plan need have any sort of compact (polynomial-size) representation.

These results are not directly applicable to the problem of finding good plans because they place no restrictions on the form of valid plans. It is possible, for example, that for a given planning domain, the only valid plans require exponential space (and exponential time) to write down. Knowing whether or not such plans exist is simply not very important.

In the present paper, we consider the complexity of a more practical and realistic problem—that of determining whether or not a plan exists in a given restricted form. The plans we consider take several possible forms that have been used in previous planning work: totally ordered plans, partially ordered plans, conditional plans, and looping plans. In all cases, we limit our attention to plans that can be expressed in size bounded by a polynomial in the size of the specification of the problem. This way, once we determine that such a plan exists, we have some hope that we can write it down in a reasonable amount of time.

In the deterministic planning literature, several authors have addressed the computational complexity of determining whether a valid plan exists, of determining whether a plan exists of a given cost, and of finding the valid plans themselves under a variety of assumptions (Chapman 1987; Bylander 1994; Bäckström & Klein 1991). These results provide lower bounds (hardness results) for probabilistic planning, since deterministic planning is a special case. In deterministic planning, optimal plans can be represented by a simple sequence of operators (totally ordered plan). In probabilistic planning, good conditional plans will often perform better than any totally ordered plan; therefore, we need to consider the complexity of the planning process for a richer set of plan structures.

The computational problems we look at are complete for a variety of complexity classes ranging from NP to PSPACE. Two results are deserving of special men-tion; first, the problem of evaluating a totally ordered plan in a succinctly represented planning domain (as might be described by a two-stage temporal Bayes network (Boutilier, Dean, & Hanks 1995)) is PP-complete. The class PP is closely related to $\#P^1$, which has been recognized as an important complexity class in computations involving probabilistic quantities, such as the evaluation of Bayes networks (Roth 1996). Of course, probabilistic computations are central to the area of uncertainty in artificial intelligence.

Second, the problem of determining whether a good totally ordered plan exists for a succinctly represented planning domain is $NP^{PP}$-complete. Whereas the class NP can be thought of as the set of problems solvable by guessing the answer and checking it in polynomial time, the class $NP^{PP}$ can be thought of as the set of problems solvable by guessing the answer and checking it using a probabilistic polynomial-time computation. It is likely that $NP^{PP}$ characterizes many problems of interest in the area of uncertainty in artificial intelligence; this paper and earlier work (Mundhenk, Goldsmith, & Allender 1996) give initial evidence of this.

## 1.1 REPRESENTING DOMAINS

A planning domain $M = \langle \mathcal{S}, s_0, \mathcal{A}, t, \mathcal{G} \rangle$ is characterized by a finite set of states $\mathcal{S}$, a finite set of operators or actions $\mathcal{A}$, an initial state $s_0 \in \mathcal{S}$, and a set of goal states $\mathcal{G} \subseteq \mathcal{S}$. The application of an action $a$ in a state $s$ results in a probabilistic transition to a new state $s'$, according to the probability transition function $t(s, a, s')$. The objective is to choose actions to move from the initial state $s_0$ to one of the goal states with probability above some threshold $\theta^2$. The state of the system is known at all times (fully observable) and so can be used to choose the action to apply.

We are concerned with two main representations for planning domains; *flat* representations, which enumerate states explicitly, and *succinct* representations (sometimes propositional, structured, or factored representations), which view states as assignments to a set of Boolean state variables. Compared to the size of the representation, flat domains typically have a polynomial number of states and succinct domains have an exponential number (though a "bad" succinct representation can be as large as a flat one).

In the flat representation, the transition function $t$ is represented by a collection of $|\mathcal{S}| \times |\mathcal{S}|$ matrices[3]. We do not treat this representation directly; see our extended technical report (Goldsmith, Littman, & Mundhenk 1997) for details on this type of problem.

In the succinct representation, straightforward probability matrices would be huge, so the transition function must be expressed another way. In artificial intelligence, two popular succinct representations for probabilistic planning domains are probabilistic state-space operators (PSOs) (Kushmerick, Hanks, & Weld 1995) and two-stage temporal Bayes networks (2TBNs) (Boutilier, Dearden, & Goldszmidt 1995). Although these representations differ in the type of planning domains they can express naturally, they are computationally equivalent; a planning domain expressed in one representation can be converted in polynomial time to an equivalent planning domain expressed in the other with at most a polynomial increase in representation size (Littman 1997).

In this paper, we use a different succinct representation for planning domains that is more closely related to representations used in the complexity theory literature. In the *circuit representation*, the transition probabilities for an action $a$, $t(s, a, s')$, are represented by a circuit of simple logic gates that takes as input succinct representations of $s$ and $s'$ and outputs a probability value in binary representation[4].

Planning domains in the PSO and 2TBN representations can be converted to the circuit representation in polynomial time, but it is not clear how to convert a circuit to a PSO or 2TBN in polynomial time. However, this conversion can be carried out by a PP machine (the basic idea is used in the proof of Theorem 2.1), so the circuit representation is equivalent to PSOs and 2TBNs in any complexity class containing PP. Since the complexity results we report for the circuit representation are all for complexity classes at least as hard as PP, these completeness results apply to PSOs and 2TBNs as well.

## 1.2 EXAMPLE DOMAIN

To help make these ideas more concrete, consider the following simple probabilistic planning domain based on the problem of building a sand castle at the beach. There are a total of four states in the domain, de-scribed by combinations of two Boolean state variables, **moat** and **castle**. The proposition **moat** signifies that a moat has been dug in the sand and the proposition **castle** signifies that the castle has been built. In the initial state, both **moat** and **castle** are false, and the two states in which **castle** is true are goal states.

---

[1]Toda (1991) showed that $P^{\#P}=P^{PP}$, from which it follows that $NP^{\#P}=NP^{PP}$. Roughly speaking, $\#P$ is as powerful as PP if used as an oracle.

[2]It is also possible to formulate the objective as one of maximizing expected discounted reward (Boutilier, Dearden, & Goldszmidt 1995), but the two formulations are essentially polynomially equivalent (Condon 1992) (the only difficulty is that succinct domains may require discount factors exponentially close to one for equivalence to hold).

[3]We assume that the number of bits used to represent the individual probability values isn't too large.

[4]This implies that the transition probabilities have at most as many bits as the circuit representing the domain has gates. There are other circuit-based representations that can represent probabilities with an exponential number of bits (Mundhenk, Goldsmith, & Allender 1996).

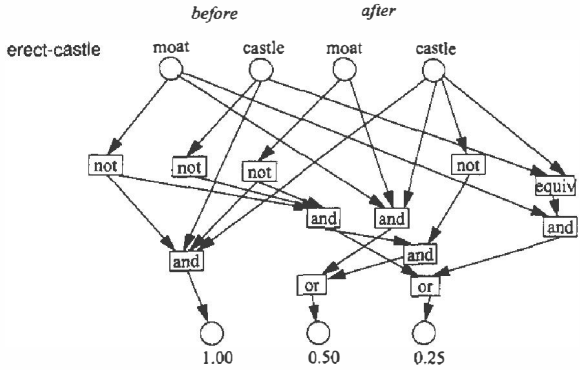

Figure 1: Circuit representation for erect-castle.

There are two actions: dig-moat and erect-castle. Executing dig-moat has two possible equiprobable effects, "no op" (state does not change), and "moat" (**moat** becomes true). The erect-castle action is more complex. If **moat** is true, then the possible effects are "castle" (probability 0.50), in which **castle** becomes true, "no op" (probability 0.25), in which the state doesn't change, and "collapse" (probability 0.25), in which **moat** becomes false. On the other hand, if **moat** is false when erect-castle is executed, then possible effects are "castle" (probability 0.25), in which **castle** becomes true, and "no op" (probability 0.75), in which the state doesn't change. The idea here is that building a moat first protects the castle from being destroyed prematurely by the ocean waves.

To illustrate the circuit representation, Figure 1 gives one possible circuit representation for the erect-castle action. This circuit takes, as input, binary representations of the "before" state $s$ and the "after" state $s'$, and outputs a binary representation of the probability of reaching $s'$ from $s$ under the erect-castle action. While this representation is not convenient for specifying complex planning domains, more natural representations can be converted into this form automatically.

### 1.3 TYPES OF PLANS

We consider four basic classes of plans for probabilistic domains: totally ordered, acyclic, looping, and partially ordered. We illustrate examples from each of these classes for the sand-castle domain in Figure 2. A *totally ordered* plan is a sequence of actions that must be executed in order. The plan terminates after the final action in the plan has been executed, or whenever a goal state is reached. For example, with probability 0.4375, the totally ordered plan in Figure 2(a)) successfully builds a sand castle.

*Acyclic* plans generalize totally ordered plans to include conditional execution of actions. They are roughly loop-free finite-state controllers for a planning domain; they express a simple type of conditional plan in which the next plan step to execute is a function of the current step and an "effect label" that describes the outcome of executing the current step. No step in an acyclic plan may be repeated more than once during plan execution. The acyclic plan in Figure 2(b) succeeds with probability 0.46875 and executes dig-moat an average of 1.75 times. Thus, it succeeds more often and with fewer actions than the totally ordered plan in Figure 2(a).

A *partially ordered plan* is a different way of generalizing a totally ordered plan. It contains no loops and no conditional branches, but can leave flexible the precise sequencing of actions (Kushmerick, Hanks, & Weld 1995). Figure 2(c) illustrates a partially ordered plan for the sand-castle domain. The dashed arrows indicate ordering constraints in contrast to solid arrows, which indicate flow of control. There are two distinct totally ordered plans consistent with the partially ordered plan in Figure 2(c): dig-moat, dig-moat, dig-moat, erect-castle, erect-castle and dig-moat, dig-moat, erect-castle, dig-moat, erect-castle.

There are several possible interpretations for how the performance of a partially ordered plan is measured. The pessimistic interpretation is that the performance of a partially ordered plan is equal to the performance of the worst possible totally ordered plan consistent with the partial order. This is closely related to the standard interpretation in deterministic partial order planning (McAllester & Rosenblitt 1991). The optimistic interpretation of the performance of a partially ordered plan is that it is the performance of the best consistent totally ordered plan, and the average interpretation is that it is the average over all possible consistent orders.

Totally ordered, partially ordered, and acyclic plans are all inherently finite horizon; plans terminate after a polynomial number of actions. *Looping* plans generalize acyclic plans to the case in which plan steps can be repeated (Smith & Williamson 1995). This type of plan is also referred to as a plan graph or policy graph (Kaelbling, Littman, & Cassandra 1995). A looping plan can express an infinite-horizon strategy because the plan will continue to execute as long as a goal state is not reached (there is no *a priori* bound on the length of the sequence of actions chosen by such a plan). For example, the looping plan in Figure 2(d) does not terminate until it succeeds in building a sand castle, which it will do with probability 1.0 eventually.

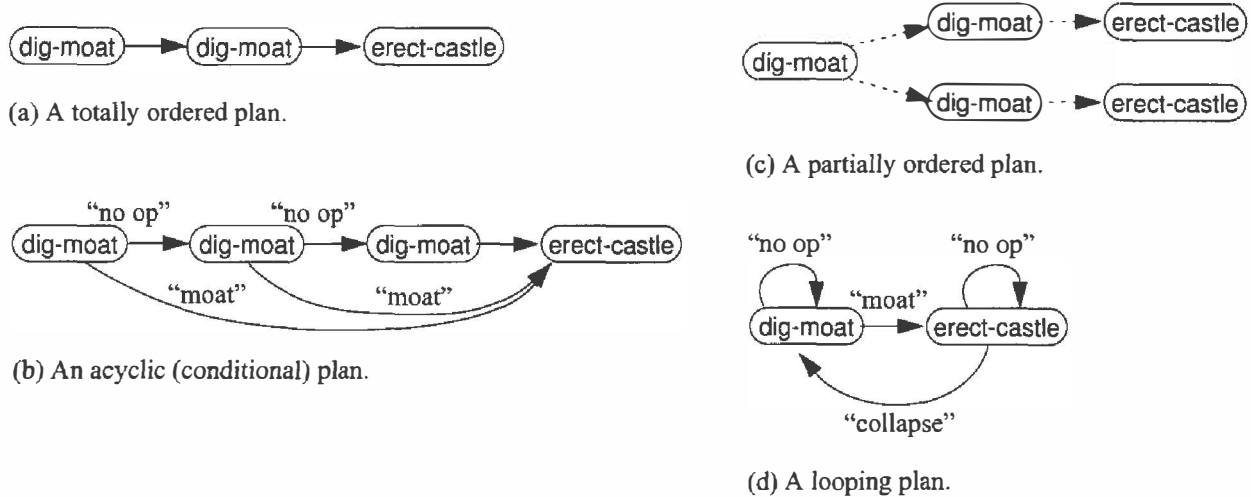

Figure 2: Example plans for the sand-castle domain.

### 1.4 DECISION PROBLEMS

Given a particular class of plans, we consider two computational problems. The first is the *plan-evaluation* problem; given a plan, a planning domain, and some threshold $\theta$, does the given plan reach the goal with probability at least $\theta$? The second problem is *plan existence*; given a planning problem and a threshold $\theta$, is there a polynomial-size plan of the required form that can reach the goal with probability at least $\theta$?

### 1.5 COMPLEXITY BACKGROUND

For definitions of complexity classes, reductions, and standard results from complexity theory, we refer to Papadimitriou's (1994) complexity textbook. In the interest of completeness, in this section we give a short description of the probabilistic and counting complexity classes we use in this work.

The class #P is the class of functions $f$ such that, for some nondeterministic polynomial-time bounded machine $N$, the number of accepting paths of $N$ on $x$ equals $f(x)$.

Probabilistic polynomial time, PP, is the class of sets $A$ for which there exists a nondeterministic polynomial time bounded machine $N$ such that $x \in A$ if and only if the number of accepting paths of $N$ on $x$ is greater than its number of rejecting paths.

For polynomial-space-bounded computations, PSPACE equals probabilistic PSPACE, and #PSPACE is the same as the class of polynomial-space-computable functions (Ladner 1989).

For any complexity classes $\mathcal{C}$ and $\mathcal{C}'$ the class $\mathcal{C}^{\mathcal{C}'}$ consists of those sets that are $\mathcal{C}$-*Turing reducible* to sets in $\mathcal{C}'$, i.e., sets that can be accepted with resource bounds specified by $\mathcal{C}$, using some problem in $\mathcal{C}'$ as a subroutine (oracle) with instantaneous output. For any class $\mathcal{C} \subseteq$ PSPACE, it is the case that $\text{NP}^{\mathcal{C}} \subseteq$ PSPACE, and therefore $\text{PSPACE}^{\text{PSPACE}}=\text{PSPACE}$; see Papadimitriou's (1994) textbook.

The complexity classes we consider satisfy the following containment properties:

$$P \subseteq \begin{array}{c} \text{NP} \\ \text{co-NP} \end{array} \subseteq \text{PP} \subseteq \begin{array}{c} \text{NP}^{\text{PP}} \\ \text{co-NP}^{\text{PP}} \end{array} \subseteq \text{PSPACE} \subseteq \text{EXP}.$$

It is known that P is properly contained in EXP.

### 1.6 SUMMARY OF RESULTS

Table 1 summarizes our results, which are explained in more detail in later sections. Table 2 summarizes a set of results for flat domains; these are described in our extended technical report (Goldsmith, Littman, & Mundhenk 1997).

## 2 ACYCLIC PLANS

Given a planning domain $M = \langle \mathcal{S}, s_0, \mathcal{A}, t, \mathcal{G} \rangle$, a plan $P = \langle \mathcal{Q}, q_0, \Sigma, \delta, \pi, \omega \rangle$ is an acyclic plan where

- $\mathcal{Q}$ and $\Sigma$ are finite sets of *plan steps* and *effects labels*, respectively,

- $q_0 \in \mathcal{Q}$ is the *initial plan step*,

186    Goldsmith, Littman, and Mundhenk

Table 1: Complexity results for succinct representations.

|  | plan evaluation | plan existence | reference |
| --- | --- | --- | --- |
| unrestricted | — | EXP-complete | Littman (1997) |
| polynomial-depth | — | PSPACE-complete | Littman (1997) |
| looping | PSPACE-complete | PSPACE-complete | Section 3 |
| acyclic | PP-complete | NP$^{\text{PP}}$-complete | Section 2 |
| totally ordered | PP-complete | NP$^{\text{PP}}$-complete | Section 2 |
| partially ordered (optimistic) | NP$^{\text{PP}}$-complete | NP$^{\text{PP}}$-complete | Section 4 |
| partially ordered (average) | PP-complete | NP$^{\text{PP}}$-complete | Section 4 |
| partially ordered (pessimistic) | co-NP$^{\text{PP}}$-complete | NP$^{\text{PP}}$-complete | Section 4 |

Table 2: Complexity results for flat representations.

|  | plan evaluation | plan existence |
| --- | --- | --- |
| unrestricted | — | P-complete |
| polynomial-depth | — | P-complete |
| looping | PL-complete | P-complete |
| acyclic | PL-complete | P-complete |
| totally ordered | PL-complete | NP-complete |
| partially ordered (optimistic) | NP-complete | NP-complete |
| partially ordered (average) | (co-)NP-hard, in PP | NP-complete |
| partially ordered (pessimistic) | co-NP-complete | NP-complete |

- $\delta : \mathcal{Q} \times \Sigma \to \mathcal{Q}$ is the (cycle free) *state-transition function*,

- $\pi : \mathcal{Q} \to \mathcal{A}$ is the *action mapping* from plan steps to actions, and

- $\omega : \mathcal{S} \to \Sigma$ is the *transition mapping* from states of the planning domain to effects labels.

Note that the quantities $\mathcal{Q}$, $q_0$, $\Sigma$, and $\delta$ jointly specify a deterministic finite-state automaton. Also, $\delta$ may be a partial function since some plan steps are final steps.

Let $M$ be a planning domain and $P$ be an acyclic plan. Then $M$ *under* $P$ behaves as follows. Both $M$ and $P$ are started "in parallel" in their initial states. Both perform steps $1, 2, \ldots$. In step $i \geq 1$, let $s$ be the current state of $M$ and $q$ be the current plan step of $P$. The current action is determined by the current state $q$ of $P$ (i.e., the new state of $M$ is $s'$ with probability $t(s, \pi(q), s')$) and $P$ gets a translation of the new state $s'$ of $M$ as an effects label (i.e., the new state of $P$ is $\delta(q, \omega(s'))$). If $\delta$ is not defined on $q$, or $s'$ is a goal state, then the process stops.

Given these definitions, we can present our first complexity result.

**Theorem 2.1** *The plan-evaluation problem for acyclic and totally ordered plans is PP-complete.*

**Proof** To show PP-hardness, we give a reduction from the PP-complete problem MAJSAT: given a Boolean formula, do the majority of assignments satisfy it?

Let $\phi$ be a Boolean formula with $n$ variables. Define the planning domain $M(\phi)$ with states $s_0, \{0,1\}^n, s_{\text{acc}}, s_{\text{rej}}$, one action $a$ and transition probabilities $t(s_0, a, w) = 2^{-n}$, $t(w, a, s_{\text{acc}}) = 1$ if $w$ satisfies $\phi$, and $t(w, a, s_{\text{rej}}) = 1$ if $w$ does not satisfy $\phi$, for $w \in \{0,1\}^n$. Let $s_{\text{acc}}$ be the only goal state. It is clear that $\phi$ is in MAJSAT if and only if $M(\phi)$ reaches the goal state with probability at least $1/2$ under the plan that repeats action $a$ twice.

For membership in PP, note that a planning domain $M$ and an acyclic plan $P$ induce a tree consisting of all paths through $M$ under $P$. This tree can be normalized in a way that makes each path have equal probability, and an accepting leaf is reached with probability at least $1/2$ if and only if $M$ reaches a goal state with probability at least $\theta$. Finally, we can define a polynomial-time probabilistic Turing machine that has this tree as its computation tree.    ∎

The plan-existence problem is essentially equivalent to guessing and evaluating a good plan, hence the problem is in NP$^{\text{PP}}$. Hardness for NP$^{\text{PP}}$ can be shown using the techniques from a paper by Mundhenk, Goldsmith, and Allender (1996). The proof uses the idea that every NP$^{\text{PP}}$ computation can be reduced to the problem of whether a succinctly described set of exponentially many plan-evaluation problems contains one that is satisfied.

**Theorem 2.2** *The plan-existence problem for acyclic and totally ordered plans is* $NP^{PP}$-*complete.*

In the above results, we consider succinctly represented planning domains but only flat plans. Succinctly represented plans are also quite useful. A *succinct acyclic plan* is an acyclic plan in which the names of the plan steps are encoded in binary and a polynomial-size circuit represents the state-transition function $\delta$. In addition, we require that the plan is at most polynomially deep even though the total number of steps in the plan might be exponential. Because the proof technique used in Theorem 2.1 generalizes to succinct acyclic plans, analogous complexity results apply. The same holds true for a *probabilistic acyclic plan*, which is a plan in which the state-transition function $\delta$ is probabilistic. These insights can be combined to yield the following corollary.

**Corollary 2.1** *The plan-evaluation problem for succinct probabilistic acyclic plans is* PP-*complete and the plan-existence problem for succinct probabilistic acyclic plans is* $NP^{PP}$-*complete.*

## 3 LOOPING PLANS

To represent looping plans, we use the same notation as with acyclic plans in the previous section, but we allow the state-transition functions to loop; this way, looping plans can be applied to infinite-horizon control. For looping plans, the complexity of plan existence and plan evaluation is quite different from the acyclic case. Looping plan evaluation is very hard.

**Theorem 3.1** *The plan-evaluation problem for looping plans is* PSPACE-*complete.*

**Proof** The plan-evaluation problem for flat domains is in PL. For a planning domain with $c^n$ states and a representation of size $n$, a looping plan can be evaluated in probabilistic space $O(\log(c^n))$, which is to say probabilistic space polynomial in the size of the input. Since probabilistic PSPACE equals PSPACE, this shows that the plan-evaluation problem for looping plans in succinct domains is in PSPACE.

It remains to show PSPACE-hardness. Let $N$ be a deterministic polynomial-space-bounded Turing machine. For any input $x$, construct a planning domain $M(x)$ that has as states all configurations of $N$ on input $x$, only one action, and state transitions with probability 1 according to the configuration transitions of $N$. All accepting configurations reach goal states. This planning domain can be encoded succinctly, and this encoding can be produced from $N$ and $x$ in polynomial time. Given a description of $N$ and $x$, one can, in time polynomial in the size of the descriptions of $N$ and $x$, produce a description of a Turing machine $N'$ that computes the transition function for $N$. In other words, $N'$ on input $c$, a configuration of $N$ (and $a$, the unique action), outputs the next configuration of $N$. (In fact, $N'$ can even check whether $c$ is a valid configuration in the computation of $N(x)$, by simulating that computation.) Because all transitions are deterministic and only one action can be chosen, it follows that the goal state is reached with probability 1 under the "constant plan" (which repeatedly chooses the only action) if and only if $N$ on input $x$ accepts. ∎

Looping plan existence is not actually any harder than looping plan evaluation, although it is still quite hard.

**Theorem 3.2** *The plan-existence problem for looping plans is* PSPACE-*complete.*

**Proof** Hardness for PSPACE follows from the same construction as in the proof of Theorem 3.1: either the "constant plan" is fine, or it is not. No other plan yields a better result.

The problem is in PSPACE because the plan being sought is no larger than the size of the succinct description of the planning domain. Thus, it can be guessed in polynomial time and checked in PSPACE. Because $NP^{PSPACE}$=PSPACE, the result follows. ∎

Recall that the *unrestricted* infinite-horizon plan-existence problem is EXP-complete; this shows the problem of determining unrestricted plan existence is EXP-hard only because some domains require plans that are larger than polynomial-size looping plans.

Theorem 3.2 shows that plan existence is PSPACE-complete in deterministic domains also. This is closely related to the PSPACE-completeness result of Bylander (1994); the main difference is that our theorem applies to more succinct plans (a single action in a loop) with more complex operator descriptions. Also, as the proofs above show, PSPACE-hardness is retained even in planning domains with only one action, so it is not simply the conditional aspect of plans that makes them hard to work with.

## 4 PARTIALLY ORDERED PLANS

A $k$-step partially ordered plan corresponds to a set of $k$-step totally ordered plans—all those that are consistent with the given partial order. The evaluation of a partially ordered plan can be defined to be the evaluation of the best, worst, or average member of the set of consistent totally ordered plans; these are optimistic, pessimistic, and average interpretations, respectively.

More formally, a partially ordered plan $P$ is a directed acyclic graph that has an action assigned to each node. A totally ordered plan $A = a_1, \ldots, a_k$ is consistent with $P$ if it satisfies the constraint that for all pairs of nodes $a_i, a_j$ if $a_i$ is an ancestor of $a_j$ in the partial order, then $i < j$, i.e., $a_i$ comes before $a_j$ in the totally ordered plan.

The plan-existence problem for partially ordered plans under the *optimistic interpretation* asks whether—given a domain $M$, a partially ordered plan $P$, and a threshold $\theta$—there is a totally ordered plan consistent with $P$ under which $M$ reaches a goal state with probability at least $\theta$. Under the *pessimistic interpretation*, we wish to know whether $M$ reaches a goal state with probability at least $\theta$ under *every* consistent totally ordered plan. Under the *average interpretation*, we wish to know whether $M$ reaches a goal state with probability at least $\theta$ averaged over all consistent totally ordered plans.

The plan-existence problem for partially ordered plans is identical to that for totally ordered plans. This is because a totally ordered plan is a special kind of partially ordered plan and its evaluation is unchanged under the pessimistic, optimistic, or average interpretations. Conversely, the value of a partially ordered plan under any interpretation is a lower bound on the value of the best totally ordered plan.

**Theorem 4.1** *The plan-existence problem for partially ordered plans is* $\text{NP}^{\text{PP}}$*-complete under the pessimistic, optimistic and average interpretations.*

The plan-evaluation problem for partially ordered plans is different from that of totally ordered plans. This is because a single partial order can encode an exponential-size set of totally ordered plans, and evaluating the partially ordered plan involves figuring out the best or worst member of this combinatorial set.

**Theorem 4.2** *The plan-evaluation problem for partially ordered plans is* $\text{NP}^{\text{PP}}$*-complete under the optimistic interpretation, co-*$\text{NP}^{\text{PP}}$*-complete under the pessimistic interpretation, and* PP*-complete under the average interpretation.*

The proofs of the first two of these results are closely related to the proof of Theorem 2.2. The average interpretation problem can be shown to be in PP by combining an argument showing how to average over consistent totally ordered plans with the argument in the proof of Theorem 2.1 showing how to evaluate a plan in a succinct domain in PP. PP-hardness follows trivially from Theorem 2.1, because totally ordered plans are a special case of partially ordered plans and evaluating totally ordered plans is PP-hard.

## 5  CONCLUSIONS

In this paper, we explored the computational complexity of plan evaluation and plan existence in probabilistic domains. We found that, in succinctly represented domains, restricting the form of the policies under consideration reduced the computational complexity of plan existence from EXP-complete for unrestricted plans to PSPACE-complete for polynomial-size looping plans to $\text{NP}^{\text{PP}}$-complete for polynomial-size acyclic plans.

The class $\text{NP}^{\text{PP}}$ promises to be very useful to researchers in uncertainty in artificial intelligence because it captures the type of problems resulting from choosing a good combinatorial structure and then evaluating its probabilistic behavior. This is precisely the type of problem faced by planning algorithms in probabilistic domains, and may capture important problems in other domains as well, such as constructing good Bayes networks from data.

The basic structure of our results is that if plan evaluation is complete for some class $\mathcal{C}$, then plan existence is typically $\text{NP}^{\mathcal{C}}$-complete. This same basic structure holds in deterministic domains: evaluating a totally ordered plan in a succinct domain is P-complete (for some typical representations) and determining the existence of a polynomial-size totally ordered plan is $\text{NP}^{\text{P}}$=NP-complete.

There are several significant plan representations that we did not explicitly consider in this work. However, the results we presented do provide a goal deal of insight into complexity results for other representations. For example, Draper, Hanks, & Weld (1994) devised a representation for partially ordered conditional plans for the C-BURIDAN system. In this representation, each plan step generates an *observation label* as a function of the probabilistic outcome of the step. Each step also has an associated set of context labels dictating the circumstances under which that step must be executed. A plan step is executed only if its context labels are consistent with the observation labels produced in earlier steps. This type of plan can be expressed as a succinct acyclic plan; Corollary 2.1 can be used to show that the plan-evaluation and plan-existence problems for partially ordered conditional plans in succinct domains are PP-complete and $\text{NP}^{\text{PP}}$-complete, respectively. Other important plan structures to which our results can be applied include universal plans or policies (Dearden & Boutilier 1997) and parallel plans (Blum & Furst 1997).

Notice that the results presented here also apply to partially observable domains (Draper, Hanks, & Weld 1994; Kaelbling, Littman, & Cassandra 1995); once we limit our decision making to following finite-state plans, it matters very little whether the true state of the world is observable or not. In many cases, the complexity of optimally solving partially observable Markov decision processes (Papadimitriou & Tsitsiklis 1987) is much higher than that of searching for a restricted controller or plan, so there is some hope of building effective algorithms based on these ideas.

The results in this paper support the intuition that searching for small plans is more efficient than searching for arbitrarily complicated plans. From a pragmatic standpoint, this suggests that exact dynamic-programming algorithms, which are so successful in flat domains, may not be as effective in succinct domains; they do not focus their efforts on the set of small plans. Algorithm development energy, therefore, might fruitfully be spent devising heuristics for problems in the class $NP^{PP}$, as this class captures the essence of searching for small plans for probabilistic domains. Heuristics for $NP^{PP}$ could lead to powerful methods for solving a range of important uncertainty-sensitive combinatorial problems.

## Acknowledgments

We gratefully acknowledge Andrew Klapper and Anne Condon for helpful conversations on this topic.